\def\year{2018}\relax
\documentclass[letterpaper]{article} %DO NOT CHANGE THIS
\usepackage{aaai18}  %Required
\usepackage{times}  %Required
\usepackage{helvet}  %Required
\usepackage{courier}  %Required
\usepackage{url}  %Required
\usepackage{graphicx}  %Required
\usepackage{color}
\newcommand{\rpm}{\raisebox{.2ex}{$\scriptstyle\pm$}}
\begin{document}
% The file aaai.sty is the style file for AAAI Press 
% proceedings, working notes, and technical reports.
%
\title{Player Experience Extraction from Gameplay Video}
\author{Zijin Luo, Matthew Guzdial, Nicholas Liao, and Mark Riedl\\
\\
School of Interactive Computing\\
Georgia Institute of Technology\\
Atlanta, GA 30332 USA\\
\{zijinluo, mguzdial3, nliao7\}@gatech.edu, riedl@cc.gatech.edu\\
}
\maketitle
\begin{abstract}
The ability to extract the sequence of game events for a given player's play-through has traditionally required access to the game's engine or source code. This serves as a barrier to researchers, developers, and hobbyists who might otherwise benefit from these game logs. In this paper we present two approaches to derive game logs from game video via convolutional neural networks and transfer learning. We evaluate the approaches in a Super Mario Bros. clone, \textit{Mega Man} and \textit{Skyrim}. Our results demonstrate our approach outperforms random forest and other transfer baselines.

\end{abstract}

\section{Introduction}
% Zijin

Analyzing gameplay is a core concern in the video game industry. Game developers can evaluate players’ experience to help improve game quality. Narrators can present better game commentary by incorporating detailed information. Tournament organizers can utilize the information extracted from gameplay to enhance the viewing experience. Researchers can analyze and compare gameplay to better understand game development and design. However, analyzing gameplay requires access to high quality representations of player experience. 
Game logs are sequences of player actions and events generated by the game engine and are the standard for representing player experience. However, access to game logs is restricted by technical issues and privacy concerns. Typically only the game's developers can access game logs. Beyond access, game logs are cumbersome given that they require patching the game for any update or change. Further, they cannot be applied outside of controlled instances, for example they cannot help in analyzing the gameplay footage of streamers or be used to understand player behavior for modded or player-created content.

Machine Learning provides one potential solution for this problem. In particular, there exists a large area of work concerned with action recognition for real world video \cite{soomro2012ucf101}, which one could imagine applying to the problem of extracting representations of player experience from gameplay video. However, such approaches require large amounts of paired training data of video labeled with activities, which restricts this approach to those with the access or resources to create a training dataset. 

In this paper, we present two approaches using convolutional neural networks (CNNs) and transfer learning to learn models that convert from gameplay video to logs. Our first model uses a CNN to predict the logs from corresponding frames in the video using a paired dataset. This model learns a set of features to track automatically, cutting back on developer authoring burden, but still requires a large dataset. The second approach consists of two parts: an existing video frame to activity model and a transfer algorithm to adapt the model to recognize and report actions in a new game domain. 

The remainder of this paper is organized as follows: first, we cover related work in approximating logs of game events or activities. Second, we overview the na\"ive CNN approach, trained on a dataset of game video and associated game events, and an evaluation of this system. Third, we cover the transfer learning approach and evaluate it compared to other transfer learning methods. We end with discussion, future work and conclusions. Our primary contributions are the application and evaluation of deep neural networks and transfer learning to the problem of player experience extraction from gameplay video. As a secondary contribution we make available a public dataset of player actions for the game \textit{Skyrim}. To the best of our knowledge this represents the first attempt to apply machine learning to the problem of deriving measures of player experience from video.

\begin{figure*}[tb]
	\includegraphics[width=\linewidth]{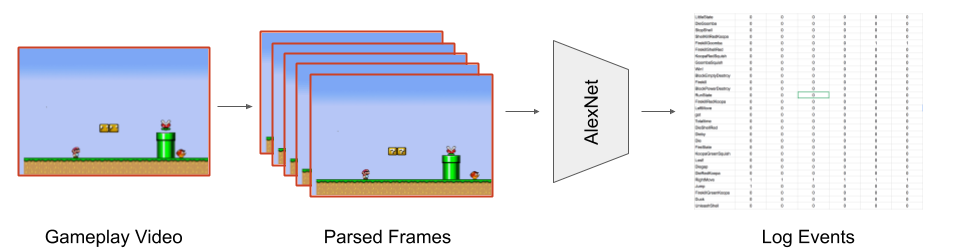}
	\caption{The mapping of gameplay video to its log events by using the na\"ive approach.}
	\label{fig:naive}
\end{figure*}

\section{Related Work}

Player modeling \cite{yannakakis2013player}, the study of computationally modeling the players of games, represents a related field to this research given that it requires representations of player experience. However, most player modeling approaches begin with the assumption that one has access to logs of game events. Towards this end prior player modeling research into Super Mario has relied on the Infinite Mario engine \cite{togelius20102009} or recreated unique Mario clones \cite{liao2017deep} to collect game logs. Alternatively, player modeling research has required a partnership or public sharing of game logs by game companies \cite{drachen2009player,sabik2015data}. In the worst case in terms of researcher effort, player modeling has required parsing gameplay videos by hand to extract events \cite{hsieh2008building}. Camilleri et al. \shortcite{camilleri2017towards} look to build a general model of player affect across different games, but this still requires access to the unique event logging system of each game. 

There exist prior approaches to extract design knowledge via machine learning for procedural content generation \cite{summerville2017procedural} that makes use of gameplay video as a data source \cite{guzdial2016game}. Summerville et al. \shortcite{summerville2016learning} extracted player paths from gameplay video, but do not extract any non-movement related events. Guzdial et al. \shortcite{guzdial2017game} learn game rules from gameplay video, which as a secondary effect outputs sequences of game events. Both approaches make use of OpenCV \cite{bradski2000opencv} to transform from raw pixels of individual frames of a video into a list of game entities and their positions on screen. However, this process requires that users provide a definition of all possible game entities to the system, which must be redefined for every new game. 

There exists a large set of applications of activity recognition for real world games and sports \cite{zhu2006player}. However, this is largely enabled by the existence of large datasets of real world human activity \cite{soomro2012ucf101}. Na\"ively, one might think that differences in game aesthetics would indicate a need for individual datasets of this size for each game to apply these methods. However, we demonstrate that one can create high-quality activity recognition models for realistic games with only a small corpus.  

To the best of our knowledge this work represents the only general approach for training automatic models for deriving player experience from gameplay video. Jacob et al. \shortcite{jacob2014non} represents the most related example of prior work. The researchers made use of a game-dependent image recognition method to collect logs of actions from gameplay video of \textit{Super Mario World}. The authors do not report results of this process, which would take substantial re-authoring to adapt to another game. Bao et al \shortcite{bao2017extracting} present an approach utilizing OpenCV to extract the location of graphical user interface windows. As previously discussed, applying OpenCV to parse video requires defining all possible entities that might be in the video, which makes generality challenging. Fulda et al. \shortcite{fulda18threat} presented an approach to derive player activity labels from gameplay video of \textit{Skyrim} using off the shelf computer vision methods to get captions of current frames and then applied natural language processing methods to classify the activity. However, they presented inconclusive results.

\section{Paired Approach}
In this section we present what one might consider the standard (or na\"ive) solution to the problem of learning a mapping of game video frames to events with convolutional neural networks (CNNs). Namely, collecting a dataset of the frames of gameplay video paired with the active game events for each frame. 

Our process is as follows. First, we collect some number of example gameplay videos. Second, we parse this gameplay video into individual frames. For the purposes of this work, we extracted 12 frames per second, as we wanted to minimize the frames that represented the same in-game events. We arrived at 12 frames per second empirically, but our approach is general to any FPS. Third, we pair these frames with the labels for active game events happening in each frame and train an AlexNet neural network \cite{krizhevsky2012imagenet} to map between video frames and the game event labels. We visualize this process in Figure \ref{fig:naive}. We chose to make use of AlexNet as it is a well-known CNN architecture with high performance on low quality image classification problems with a large number of classes. AlexNet has been described in more detail in prior publications, but it has five CNN layers and three fully connected layers. 
We chose to use \textit{Double Cross Entropy} as the loss function, and \textit{Adam} as the optimizer.

To apply AlexNet to this class we alter the size of the fully connected layer to be of size $E$, where $E$ indicates the maximum number of in-game events. AlexNet's final fully connected layer makes use of ReLU activation, which means each output varies from -1.0 to 1.0. For predicting what final events occurred in a given frame we make use of a threshold of 0.5, marking each appropriate event as active if its index in the vector exceeds this value. We settled on 0.5 after examining the final performance of the trained model before comparing it against the test set.
This final layer then becomes a classification over a multi-hot vector of all game events. For example if the only first event were active in a frame we would represent that as $\langle 1,0,0... 0\rangle$, if only the second event were active in a frame we would represent that as $\langle 0,1,0... 0\rangle$, and so on. It is possible and even likely that many events will co-occur in the same frame. 
We make use of PyTorch \cite{paszke2017automatic} for training our version of AlexNet.
\begin{figure}[tb]
	\includegraphics[width=\linewidth]{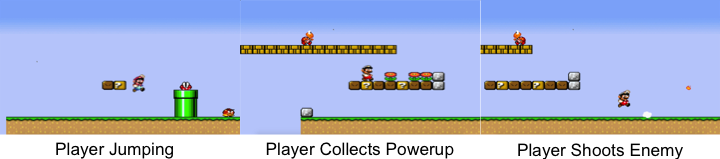}
	\caption{Examples of three frames in Gwario and the corresponding event occurring.}
	\label{fig:gwarioExamples}
\end{figure}

\section{Paired Data Approach Evaluation}

In this section we evaluate the application of our paired data approach for learning a mapping from gameplay video frames to game events. To evaluate this approach we make use of a game for which we have access to the underlying game logging system, a games with a purpose (GWAP) clone of Super Mario  Bros. called {\em Gwario} \cite{siu2017evaluating}. This logging system contains thirty possible event types such as an enemy death, a player collecting a coin, etc. We give examples with three frames with one of these event occurrences in Figure \ref{fig:gwarioExamples}. Note that these events capture both the player's behavior and non-player actions in the game.

Given that we focus on reducing the authoring burden on those who could benefit from these models we only make use of two instances of gameplay, using two gameplay videos and their associated game logs as our data for this evaluation. This represents a total of 3500 instances of frames paired with the events currently occurring in said frame. Due to the nature of Gwario, which required players to make classification decisions for images of purchasable goods, the majority of game frames had no events. This means that a na\"ive system that guessed no event occurred for each frame could achieve an accuracy of 88\%. For frames with multiple events we calculated partial accuracy for that frame as $1-x/n$, where $x$ is the number of incorrect event guesses and $n$ is the total number of event guesses.
We make use of a five fold cross-validation, which translates to roughly a minute of test video for each video, given our choice of frames per second.

We compare our approach to a random forest given its prior uses as a baseline for similar prior work \cite{guzdial2016deep}. We make use of the SciPy Random Forest implementation \cite{jones2014scipy}, which is a 10 tree random forest. To encourage generality we limit the depth to 100 layers. We further make use of a random baseline, which randomly predicts a random number of events (up to the maximum number of co-occurring events) for each frame. 

We summarize the results in terms of average test accuracy in Table 1 across all five folds. Across all folds our approach outperforms the random forest baseline by an average of roughly 10\%, only exhibiting about 5\% test error. Given that there were only thirty possible events, this indicates that on average AlexNet got one to two events incorrect for each frame. Comparatively, the random forest classifier had a test error of roughly 15\%, meaning 4-5 incorrect events per frame on average. 
Guessing no event at all outperforms the random forest baseline, which may run counter to one's expectations. This issue will be general to other games in which players are not constantly performing actions. This points to the difficulty of the problem and better situates the significance of AlexNet.

\begin{table}[t]
\caption{Comparison of the average test accuracy of a 5 fold cross-validation between our approach and three baselines.}
\label{tab:gwarioResults}
\begin{center}
\begin{tabular}{ |c|c|c|c| } 
 \hline
  AlexNet& Random Forest & Random & No-Event \\ 
  \hline
 \textbf{94.01\rpm0.68} & 85.91\rpm0.67 & 0.97\rpm0.83 & 88.0\rpm0.0\\ 
  \hline
\end{tabular}
\end{center}
\end{table}

\section{Transfer Approach}

Our paired data approach has three major issues limiting its practicality. First, it requires a large, well-labeled dataset. Second, it takes a large amount of training and computation power to train a new model. Third, it does not demonstrate a sufficient level of performance to replace within-engine logging systems, still exhibiting 5\% test error. In this section we demonstrate a second approach that uses transfer learning to address these limitations. 

Transfer learning typically takes the form of training a neural network architecture on some existing, large, and well-labeled dataset such as ImageNet \cite{deng2009imagenet}. The final layer of this neural network is then retrained to adapt the model to a novel domain, freezing the other layers to leverage the high quality features learned during the initial training. Transfer learning tends to require less training time on the new domain than training from scratch. 

A standard transfer learning approach may not suffice.
Consider the case of trying to train a frame-to-events model for Gwario by transfering a model from another, more general dataset such as ImageNet. 
ImageNet has 1000 label classes, requiring a neural network with a final layer of size 1000.
To transfer the model to Gwario, which has 30 event types, a standard transfer learning approach would replace the network's final layer with one of length 30 and retrain just the final, fully-connected layer on available Gwario log data while holding all other weights constant. However, we would not anticipate the same features from ImageNet, composed of images of real world objects and animals to work well with the pixel graphics of Gwario.

\begin{figure*}[tb]
\centering
	\includegraphics[width=6in]{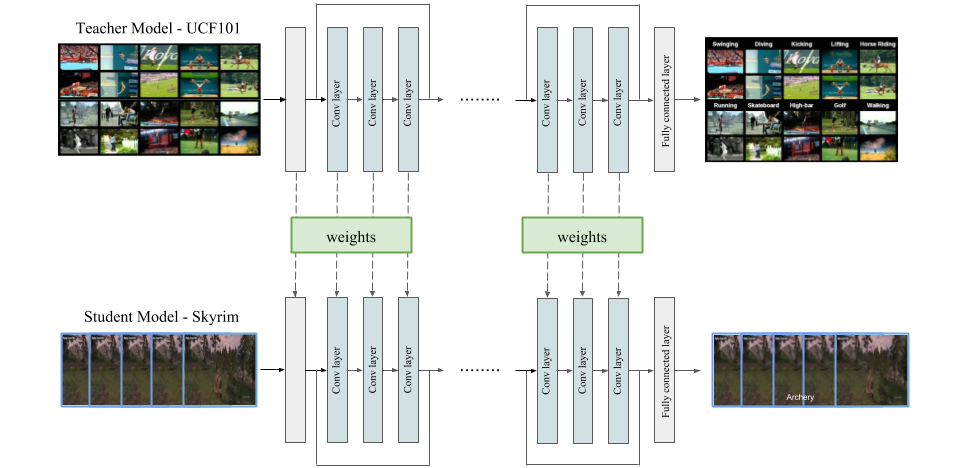}
	\caption{The mechanism of student-teacher method: creating a student model from the teacher model by copying the weights from it and retrain on the new dataset}
	\label{fig:teacher-student-figure}
\end{figure*}

To address the issue of mismatching features we use a specific transfer learning method called {\em student-teacher learning}, sometimes called {\em knowledge distillation} \cite{wong2016sequence,furlanello2018born}. The student-teacher method replicates one network's weights into an independent network with a distinct and, typically, smaller architecture. One only trains the ``teacher" model once, which can then be used as the basis for an arbitrary number of ``student" networks, which require a comparatively smaller dataset and training time. We visualize this approach in Figure \ref{fig:teacher-student-figure}. The intuition behind this approach is that you are giving the network a starting point in a related domain (e.g. \textit{Super Mario Bros.}) and then training it as usual to adjust it to a related domain (e.g. \textit{Megaman}). The difference between this and more standard transfer learning method is that the entirety of the student network is retrained after transfer, allowing the system to learn new features. Notably student-teacher approaches typically involve going from a larger to a smaller network, but this is not a requirement. 

This approach helps to overcome the limitations of the na\"ive supervised approach to log generation. 
First, it requires a much smaller dataset for the target game. 
Second, it requires a much shorter training time to converge.
Third, student-teacher networks typically demonstrate higher performance than training on the same dataset na\"ively \cite{furlanello2018born}.

\begin{figure}[tb]
	\includegraphics[width=\linewidth]{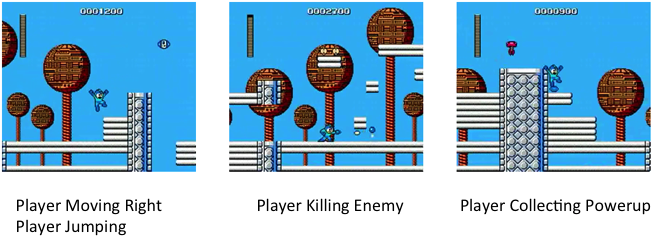}
	\caption{Examples of frames tagged with actions from the \textit{Mega Man} dataset.}
	\label{fig:megamanActions}
\end{figure}

\section{Transfer Evaluation: Mega Man}

We investigate the application of this transfer approach to a pixel-based game that parallels the na\"ive supervised data approach. We took a single gameplay video of a single level of \textit{Mega Man} from the Nintendo Entertainment System and hand-coded it with five of the thirty event types from our paired data approach evaluation. These were {\em player moving right}, {\em player moving left}, {\em player jumping}, {\em player shoots enemy}, and {\em player collects powerup}. 
We visualize four instances of these events across three frames in Figure \ref{fig:megamanActions}. We then ran an evaluation with a 80-20 train-test split, using the AlexNet model trained on Gwario from the na\"ive supervised data approach evaluation as the teacher and a new AlexNet model as the student network.

We evaluate against two baselines. For the first we compare against domain adaptation \cite{tommasi2013frustratingly}, in which backpropagration was used to train the same architecture on a combined dataset of the five shared classes in Gwario and \textit{Mega Man} videos. 
This technique is much slower but represents a na\"ive approach. 
For the second baseline we make use of the same random guessing baseline from the prior evaluation. 
Our \textit{Mega Man} video was downloaded from YouTube and represents expert play. Thus, there were few frames without action, but always guessing no event could still achieve an accuracy of roughly 74\%.

\begin{table}[tb]
\caption{Comparison of the average accuracy of our student-teacher transfer learning approach for Mega Man to three baselines.}
\label{tab:megamanResults}
\begin{center}
\begin{tabular}{ |c|c|c|c|c| } 
 \hline
  Student-Teacher & adaptation & Random & No-Event \\ 
  \hline
 \textbf{80.92\rpm0.06} & 80.09\rpm0.05 & 15.50\rpm0.02 & 73.78\rpm0.0\\ 
  \hline
\end{tabular}
\end{center}
\end{table}

We summarize the results in Table~\ref{tab:megamanResults}. The student teacher performs the best, but only roughly 0.8\% better than the domain adaptation approach. This is a small improvement over the paired data approach, but this may be due to the fact that Gwario was not similar enough in aesthetic to \textit{Mega Man} to fully harness the potential of the student-teacher approach. However, the student-teacher approach only took a single epoch to converge, making it far more time efficient than the paired data approach and domain adaptation baseline.

\begin{figure}[tb]
	\includegraphics[width=\linewidth]{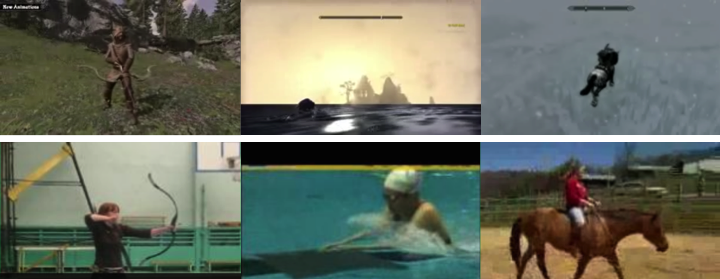}
	\caption{Comparison between three Skyrim actions (top) and UCF-101 actions (bottom). From left to right archery, breaststroke, and horse riding.}
	\label{fig:skyrimUCF}
\end{figure}

 \begin{figure}[tb]
	\includegraphics[width=\linewidth]{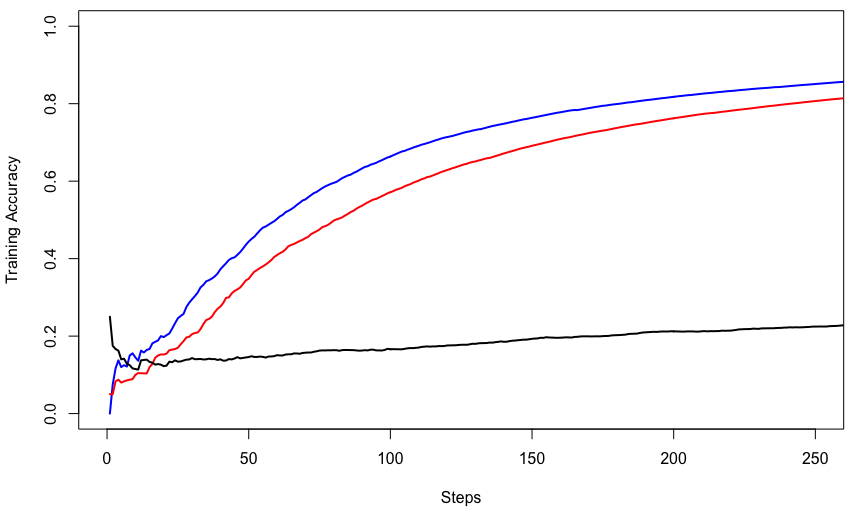}
	\caption{The training accuracy over the first 250 epochs for the student-teacher approach (blue), ImageNet transfer baseline (red), and domain adaptation baseline (black).}
	\label{fig:traininAccuracy}
\end{figure}

\section{Transfer Evaluation: Skyrim}

In the previous section we demonstrate the difficulty in transferring a model from one sprite-based game to another sprite-based game. 
That makes sense when logs are available for both games.
In this section we show that our transfer learning approach can be used to generate logs for a relatively photorealistic game---{\em Skyrim}---given a real world video dataset.
We trained a teacher network on the UCF-101 dataset \cite{soomro2012ucf101}, which is a comprehensive video dataset of 101 human behaviors. 
We made this choice given that it allows our teacher model to obtain an excellent basis for real human action recognition. 
For our teacher and student neural network architecture we make use of the resnet-152 model \cite{he2016deep}, given that it is popular and has known high-quality performance. 
We lack the space for a full description of resnet-152, but note that it is a two-stream 152-layer CNN \cite{simonyan2014two}, meaning it takes both raw pixel images and the difference between each pair of frames as input. While this is a very large neural network, it is less complex than the similarly popular VGG net \cite{simonyan2014very}. 
Further, due to its popularity it is possible to find pre-trained resnet-152 models, further reducing the computation burden.

For our transfer domain, we use the game, \textit{Skyrim}, a popular 3rd person perspective role-playing adventure game in which players can engage in a wide variety of activities. 
We further chose it as our domain for the evaluation given that there exists prior research that could benefit from a more accurate activity model of \textit{Skyrim }\cite{fulda18threat}. 

We compiled ten, five second gameplay clips of ten player activities in \textit{Skyrim} from YouTube, which we make publicly available.\footnote{https://github.com/IvoryCandy/Skyrim-Human-Actions}
These activities are: archery, breaststroke, crossbow, dance, dodge, fly, horse riding, run, skydiving, and waving weapon. These activities are also ten of the one-hundred and one activity labels present in UCF-101, which in theory should improve the transfer learning performance. 
We visualize three of these activities in Figure \ref{fig:skyrimUCF} for both UCF-101 and \textit{Skyrim}. Notably, the \textit{Skyrim} frames include graphical user interface elements, and include some modded or player-created elements.

The procedure to apply the student-teacher method was straightforward. 
We first trained our resnet-152 teacher model on UCF-101, which obtains a testing accuracy of 81\% on the held-out UCF-101 test batch. 
We then prepared another resnet-152 as the student model. 
As shown in the Figure \ref{fig:teacher-student-figure}, we copied the weights from the teacher model to the student model and re-sized the last fully connected layer to match the ten labels in our new dataset for the target game. Then, we trained the student model on the new dataset.

We make use of two baselines. For the first we compare against domain adaptation, in which backpropagration was used to train the same architecture on a combined dataset of the ten shared classes in UCF-101 and our Skyrim videos. 
As a second baseline we compare against the same architecture first trained on ImageNet and then applied transfer learning to retrain the final layer to classify for our ten cases, which represents a typical transfer learning approach. 
We ran 50-50, 66-33, and 83-17 train-test splits across our Skyrim video dataset. 
The last split is atypical---one might instead see an 80-20 split---but it fit more evenly given the size of our dataset.

\begin{table}[tb]
\caption{A comparison of our student-teacher transfer learning approach compared to an ImageNet baseline and a na\"ive domain adaptation approach.}
\label{tab:skyrimResults}
\begin{center}
\begin{tabular}{ |l|c|c|c| } 
 \hline
  & UCF-101 & ImageNet & adaptation \\ 
  \hline
 50-50 & \textbf{99.92\rpm0.08} & 99.87\rpm0.10 & 74.78\rpm18.94 \\ 
  \hline
 66-33 & \textbf{99.99\rpm0.02} & 99.94\rpm0.05 & 83.91\rpm13.04 \\ 
  \hline
 83-17 & \textbf{100.00\rpm0.02} & 99.96\rpm0.06 & 90.40\rpm13.04 \\ 
 \hline
\end{tabular}
\end{center}
\end{table}

We compare the average results with standard deviation for all three splits in Table \ref{tab:skyrimResults}. Given there were only ten possible activities and no overlap a random baseline would average around 10\% test accuracy, which all approaches outperformed. Our approach outperformed both other approaches consistently, performing near-perfectly. ImageNet transfer has similarly high accuracy, but distribution of accuracies across all test data was greater. 

Figure \ref{fig:traininAccuracy} represents the average training accuracy for every training step for all three approaches across all splits. The teacher-student network is in blue, the ImageNet baseline is in red, and the domain adaptation approach using backpropagation is in black. This demonstrates that the student-teacher approach converged faster than the ImageNet baseline. However, both approaches eventually converged to nearly 1.0 training accuracy.

Our approach for deriving player activities for realistic games shows excellent performance, even only given five training videos for each of the target activities. The ImageNet-based transfer approach also performed well, which does not benefit from the similarity in real human and humanoid 3D game characters. We take this as a positive sign, that there exists a large amount of high quality datasets one could pull from for this video-to-log task. It also demonstrates the ability for CNNs to adapt between real world features and games with realistic aesthetics quickly.

\section{Discussion}

We present two approaches in this paper for going from video to a representation of player experience. The best version of our approach in terms of performance was our student-teacher approach to \textit{Skyrim} with UCF-101 as the teacher training set. This demonstrates the importance of large, high quality datasets. Compare this to the Gwario evaluation and Mega Man evaluations, which drew on only two videos as their dataset or teacher network training set. We anticipate that the datasets one has access to will have a major impact in the success of these approaches. In particular, we anticipate the higher performance of \textit{Skyrim} was due to the fact that general, real world features translated to photo-realistic 3rd person games better than sprite-based games where game art is highly stylized. 

We acknowledge that such deep neural network approaches to function approximation for gameplay video to player experience will never be as accurate as access to a game engine logging system of game events. However, the results of this paper indicate that it can be a viable alternative in cases in which one has the resources to tag gameplay video as in the first approach or an existing trained model and some smaller tagged dataset as in the second approach. However, we do not advocate for the use of either approach in parsing single videos of a specific player experience, given that there will be some amount of error or noise. Instead we anticipate applying this approach when a researcher, hobbyist, or developer wishes to parse a large amount of gameplay video to extract overall trends, or other cases where the error rate's impact has less impact.

We note that training an initial teacher model can be computationally intensive. We found that our resnet-152 teacher model for both UCF-101 and ImageNet took several days to initially converge, even training on high-powered GPUs. We anticipate researchers will have more success finding pre-trained models and using these as a teacher model, given that training the student network can take as little as a single epoch. However, this is still a limitation on the current work given our desire to make parsing gameplay video to extract game events and player activities more accessible. 

\section{Future Work}

Our results prove that we can develop video action recognition models for some video games. However, our transfer learning approach relies on the existence of an existing, high-quality dataset or a pre-trained model on such a dataset. This is a limitation, especially when it comes to visual aesthetics. The majority of high-quality datasets in existence for image recognition are composed of realistic images, which limits the potential application of these datasets to games with a pixel art, cartoon-like, or generally unrealistic visual aesthetics, given that we anticipate very different features. Even with transfer learning, there is still a need for powerful computation to train the initial model. For future work we intend to explore the possibility of stylistically altering these realistic datasets to particular game aesthetics.

We anticipate even with the performance of our current techniques that there are many possible applications. For example, we imagine applying these approaches as a means of generating features for game human subject studies in which one does not have access to the underlying game engine as in \cite{fulda18threat}. Alternatively, we anticipate that such a system could be applied to extract player experience measures for applications into player modeling, for example determining categories of player types for new games or genres \cite{yannakakis2013player}. Further, given the speed of a neural network at test time, we anticipate the ability to apply this system to applications of live game commentary or shout-casting, either human or artificially intelligent \cite{dodge2018experts}.

\section{Conclusions}

In this paper we present two approaches for learning a model that converts from gameplay video to representations of player experience. Our first approach represents a standard application of convolutional neural networks to this problem, training on a dataset of gameplay video frames paired with game events. Our second approach to derives a model through transfer learning on small sets of tagged videos. We evaluate these models for a Super Mario clone, \textit{Mega man} and \textit{Skyrim}, and present a corpus we developed for the latter. Our results demonstrate that both approaches stand as reasonable approximations of true game logs. We hope that researchers may apply these methods to further all areas of games research.

\section{Acknowledgements}

This material is based upon work supported by the National Science Foundation under Grant No. IIS-1525967. Any opinions, findings, and conclusions or recommendations expressed in this material are those of the author(s) and do
not necessarily reflect the views of the National Science Foundation.

\bibliographystyle{aaai}
\bibliography{aaai}

\end{document}